\documentclass[sigconf,nonacm]{acmart}
\AtBeginDocument{%
  \providecommand\BibTeX{{%
    \normalfont B\kern-0.5em{\scshape i\kern-0.25em b}\kern-0.8em\TeX}}}

\usepackage{multirow}
\usepackage{subcaption}
\usepackage{amsfonts,mathtools}
\usepackage{subfloat}
\usepackage{algorithm,algpseudocode}
\usepackage{adjustbox}
\usepackage{makecell}

\usepackage{tcolorbox}
\tcbuselibrary{listingsutf8}

\usepackage{xcolor}

\newtcolorbox{generationbox}[1][]{colback=blue!5!white, colframe=blue!75!black,
  fonttitle=\bfseries, title=, listing only,
  listing options={basicstyle=\ttfamily\footnotesize, breaklines=true, columns=fullflexible},
  #1}

\newtcolorbox{fafinetuningbox}[1][]{colback=red!5!white, colframe=red!75!black,
  fonttitle=\bfseries, title=, listing only,
  listing options={basicstyle=\ttfamily\footnotesize, breaklines=true, columns=fullflexible},
  #1}

\newtcolorbox{dufinetuningbox}[1][]{colback=green!5!white, colframe=green!75!black,
  fonttitle=\bfseries, title=, listing only,
  listing options={basicstyle=\ttfamily\footnotesize, breaklines=true, columns=fullflexible},
  #1}

\usepackage{multirow}
\usepackage{subcaption}
\usepackage{amsfonts,mathtools}
\usepackage{hyperref}
\usepackage{url}
\usepackage[normalem]{ulem}
\useunder{\uline}{\ul}{}
\usepackage{fixmath}

\begin{document}
%\fancyhead{}
\title{TeleDoCTR: Domain-Specific and Contextual Troubleshooting for Telecommunications}

%\iffalse
\author{Mohamed	Trabelsi}
\email{mohamed.trabelsi@nokia-bell-labs.com}
\affiliation{%
 \institution{Nokia Bell Labs}
  %\streetaddress{113 Research Drive (Building C)}
  \city{Murray Hill}
  \state{NJ}
  \country{USA}
  %\postcode{18015}
}

\author{Huseyin Uzunalioglu}
\email{huseyin.uzunalioglu@nokia-bell-labs.com}
\affiliation{%
  \institution{Nokia Bell Labs}
  %\streetaddress{113 Research Drive (Building C)}
  \city{Westford}
  \state{MA}
  \country{USA}
  %\postcode{18015}
}
%\fi
%\renewcommand{\shortauthors}{Trabelsi et al.}

\begin{abstract}

Ticket troubleshooting refers to the process of analyzing and resolving problems that are reported through a ticketing system. In large organizations offering a wide range of services, this task is highly complex due to the diversity of submitted tickets and the need for specialized domain knowledge. In particular, troubleshooting in telecommunications (telecom) is a very time-consuming task as it requires experts to interpret ticket content, consult documentation, and search historical records to identify appropriate resolutions. This human-intensive approach not only delays issue resolution but also hinders overall operational efficiency. To enhance the effectiveness and efficiency of ticket troubleshooting in telecom, we propose TeleDoCTR, a novel telecom-related, domain-specific, and contextual troubleshooting system tailored for end-to-end ticket resolution in telecom. TeleDoCTR integrates both domain-specific ranking and generative models to automate key steps of the troubleshooting workflow which are: routing tickets to the appropriate expert team responsible for resolving the ticket (classification task), retrieving contextually and semantically similar historical tickets (retrieval task), and generating a detailed fault analysis report outlining the issue, root cause, and potential solutions (generation task). We evaluate TeleDoCTR on a real-world dataset from a telecom infrastructure and demonstrate that it achieves superior performance over existing state-of-the-art methods, significantly enhancing the accuracy and efficiency of the troubleshooting process.

\end{abstract}

\begin{CCSXML}
<ccs2012>
   <concept>
       <concept_id>10010147.10010178.10010179.10010182</concept_id>
       <concept_desc>Computing methodologies~Natural language generation</concept_desc>
       <concept_significance>500</concept_significance>
       </concept>
   <concept>
       <concept_id>10010147.10010257.10010258.10010259.10003268</concept_id>
       <concept_desc>Computing methodologies~Ranking</concept_desc>
       <concept_significance>500</concept_significance>
       </concept>
   <concept>
       <concept_id>10010147.10010257.10010282.10010290</concept_id>
       <concept_desc>Computing methodologies~Learning from demonstrations</concept_desc>
       <concept_significance>500</concept_significance>
       </concept>
 </ccs2012>
\end{CCSXML}

\ccsdesc[500]{Computing methodologies~Natural language generation}
\ccsdesc[500]{Computing methodologies~Ranking}
\ccsdesc[500]{Computing methodologies~Learning from demonstrations}

%\fi

\keywords
{ticket troubleshooting, large language models, retrieval models, instruction-tuning}

\maketitle

\section{Introduction}

Service management systems are widely used across industries to ensure the reliable delivery and operation of complex services. A notable example is the telecommunications (telecom) sector, where services such as mobile voice, messaging, data, and home internet are provided through complex infrastructures that rely on wireless, optical, and internet protocol (IP)-based networks. Analogous service management systems are prevalent in other domains as well, including information technology, industrial automation, energy, and transportation. These service management systems typically consist of two main components: (1) monitoring and logging systems that continuously collect health and performance data from the service infrastructure, and (2) diagnostic tools that analyze the monitoring data to infer the health of the system, detect anomalies, and identify potential faults. Historically, the process of resolving service problems, known as ticket troubleshooting, has relied heavily on human expertise. When a ticket is created, it generally includes a brief title, a natural-language description of the problem, and metadata specifying the affected system components. Engineers then retrieve and analyze system diagnostics and logs to investigate the issue and determine a resolution strategy. Ticket troubleshooting is a critical operation, as it directly impacts customer satisfaction and service reliability. However, it remains a time-intensive process, not only due to the complexity of resolving each issue, but also because of the delays it introduces to other engineering tasks. 

Many existing methods in ticket troubleshooting \cite{feng2022tadaa,liu2023ticket,bosch2022fine,sample2018predicting,ferland2020automatically} focus on the ticket classification task in terms of assigned class, resolution class, or resolution time as part of the ticket troubleshooting process. Other methods \cite{cristian2019study,bosch2022fine} aim to identify duplicate or contextually similar tickets to aid in problem resolution. Some techniques also provide solution recommendations based on historical data \cite{ferland2020automatically,grimalt2022berticsson}. Additionally, large language model (LLM)-based solutions \cite{arici2023llm,JainGN24,RoyZBBLFR24} have been introduced to offer real-time, conversational assistance for troubleshooting common technical issues. While effective for frequently asked questions and general-purpose support, these LLMs often fall short in specialized domains such as telecom, where domain-specific knowledge, expertise, and reasoning are essential. Moreover, most existing methods address only a single component of the troubleshooting process, classification, retrieval, or generation, without offering an integrated and unified troubleshooting solution. This fragmented approach creates a significant burden when attempting to build end-to-end systems that can fully automate and support the ticket troubleshooting workflow in complex and domain-specific environments.

Inspired by the recent progress of domain-specific LLMs 
%in several domains 
\cite{BloombergGPT,WizardMath,AlphaGeometry,MedPaLM-2,ChatLaw}, we introduce a novel \textit{\textbf{Tele}com-related} \textit{\textbf{Do}main-specific} and \textit{\textbf{C}ontextual} \textit{\textbf{TR}oubleshooting} (\textbf{\textit{TeleDoCTR}}) system tailored for end-to-end ticket resolution in telecom. TeleDoCTR integrates both domain-specific ranking and generative models to automate the full troubleshooting process composed of routing tickets to the appropriate expert team responsible for resolving the given ticket, retrieving contextually and semantically similar historical tickets to the given ticket for broader contextual information incorporation, and finally generating a detailed fault analysis report outlining the issue, root cause, and potential solutions. The TeleDoCTR architecture comprises four key components: (1) domain-specific rankers for ticket similarity and generated fault analysis report evaluation, (2) a classification module for predicting the responsible resolution team of the ticket, (3) a fault analysis generation module using finetuning, multi-response generation, and multi-response ranking, and (4) a Retrieval-Augmented Generation (RAG) conversational module enhanced with demonstrations selection. To enable efficient domain adaptation, TeleDoCTR employs Parameter-Efficient Fine-Tuning (PEFT) via QLoRA \cite{hu2022lora,DettmersPHZ23} for both ticket classification and fault analysis generation. The generative model produces multiple candidate fault analysis reports, and finetuned domain-specific rankers evaluate their semantic relevance to the input ticket. The top-ranked reports are then returned to the user. Additionally, the domain-specific rankers are used to select relevant demonstrations for the RAG module, enabling more accurate and context-aware multi-turn fault analysis generation. We train and evaluate TeleDoCTR using a large-scale telecom-related troubleshooting dataset that reflects the inherent complexity and diversity of telecom troubleshooting scenarios. Experimental results show that TeleDoCTR significantly outperforms general-purpose LLMs and isolated solutions, delivering more effective and efficient ticket resolution in real-world telecom environments.

In summary, we make the following contributions: 

%\iffalse
\begin{itemize}
    \item We propose a new telecom-related, domain-specific, and contextual troubleshooting system (TeleDoCTR) that combines domain-adapted ranking and generative models to automate the three key steps of ticket troubleshooting in a unified and integrated solution.
    \item We finetune specialized ranking models on the troubleshooting dataset to measure semantic relevance for ticket similarity and fault analysis evaluation. We also perform instruction-tuning using QLoRA to finetune a domain-specific LLM for both ticket classification and fault analysis generation. 
    \item We develop a response-ranking mechanism in which multiple fault analysis reports are generated by the finetuned LLM and ranked using the domain-specific rankers to identify the most relevant fault analysis reports.
    \item We leverage the domain-specific ranking models to select demonstrations (ticket-fault analysis pairs), which are incorporated into a RAG conversational module for improved multi-turn fault analysis generation.
    \item We use large-scale telecom-based troubleshooting data for training and evaluation, and demonstrate that our new system outperforms baselines in all three key steps. 

\end{itemize}

\section{Related work} 

\subsection{Ticket Troubleshooting}

Ticket troubleshooting involves analyzing and resolving issues submitted through a ticketing system, which is a critical process in industries such as telecom, information technology, industrial automation, energy, and transportation. Existing methods \cite{feng2022tadaa,arici2023llm,liu2023ticket,bosch2022fine,sample2018predicting,ferland2020automatically} have primarily focused on classification tasks, such as predicting the team responsible for resolving the ticket, the resolution category, or the expected resolution time, as part of the broader troubleshooting pipeline. While these approaches offer value in streamlining ticket routing, they fall short in covering the interpretability and depth that are required for complex issue resolution. Our system advances beyond the traditional classification by generating detailed natural language fault analysis reports. These reports explicitly describe the identified issue, its root cause, and resolution steps. This shift from assigning discrete labels to producing human-readable outputs enhances the usability and trustworthiness of the troubleshooting, and narrows the gap between machine-generated responses and human-level understanding.

Several existing approaches \cite{cristian2019study,bosch2022fine} have explored the ticket retrieval task by applying traditional information retrieval techniques and semantic similarity models to identify previously resolved tickets or recommend related troubleshooting cases. The primary goal of these methods is to accelerate the resolution process by retrieving historical examples that are contextually or semantically similar to the new ticket, which enables engineers to reuse past knowledge and solutions. Our system takes this paradigm a step further by embedding an interpretive layer within the retrieval process. Rather than merely surfacing similar tickets, our system leverages LLMs to actively analyze and contextualize the retrieved information in relation to the current issue. This includes identifying similarities, extracting relevant diagnostic patterns, and inferring how previous fault analysis reports can inform the resolution of the present ticket. By drawing these explicit connections between historical cases and the current ticket, our system transforms retrieval from a passive reference tool into an active decision-support mechanism.

LLMs have been explored in the context of troubleshooting \cite{arici2023llm,JainGN24,RoyZBBLFR24}, primarily aiming to provide users with real-time, conversational support for resolving common technical issues. These systems are generally effective for handling frequently asked questions and offering assistance based on general-purpose knowledge. However, their utility significantly decreases in specialized domains such as telecom, where troubleshooting requires deep domain expertise, complex reasoning, and the ability to interpret context-specific data. The general-purpose LLMs, that are trained on broad knowledge bases, lack the contextual awareness needed to address the various technical issues found in telecom infrastructure and operations. Our solution incorporates domain-specific finetuned models using real-world troubleshooting data, which allows the system to align closely with the domain-specific knowledge of troubleshooting.

\subsection{Domain-specific LLMs}

Deep contextualized language models, such as BERT \cite{Devlin2019BERTPO} and RoBERTa \cite{Liu2019RoBERTaAR}, have been proposed to solve multiple tasks in information retrieval (IR) \cite{sakata2019,dai2019,strubert,Chen2020TableSU,survey_doc_retrieval,khattab_sigir,nogueira_passage,nogueira_multi} and natural language processing (NLP) \cite{WangHCS20,dame,selab_arxiv,selab,WaddenWLH19,absformer}. Recently, researchers have focused on the Generative Pretrained Transformer (GPT) models to advance LLM capabilities in multiple tasks \cite{SunL0WGZ023,WangYW24,zhang2023,MinRSVNSAHR24,abs-2209-12356}. In this regard, multiple general-purpose LLMs, such as ChatGPT, LLaMA, and Mistral, have been developed to generalize to multiple tasks. Although general-purpose LLMs demonstrate remarkable performance across a wide range of tasks, their effectiveness diminishes substantially in domains that demand deep and specialized knowledge, such as mathematics, finance, healthcare, and telecom. This limitation arises from the nature of their training data, which typically consists of large-scale heterogeneous text sources. As a result, these models often lack the precision and contextual understanding required for domain-specific applications. To overcome this challenge, domain-specific LLMs have emerged as a promising solution. By finetuning general-purpose models on curated domain-specific datasets, these specialized models can be optimized to capture the technical language and reasoning patterns that are unique to a given domain. This specification significantly enhances the LLM's performance and reliability in expert-level scenarios.

A growing number of domain-specific LLMs have been proposed to address the limitations of general-purpose LLMs in tasks requiring deep domain-specific expertise. For instance, BloombergGPT \cite{BloombergGPT} is trained exclusively on financial data to enhance understanding and generation of texts in the financial domain. In the field of mathematics, WizardMath \cite{WizardMath} is developed to address complex mathematical problems that general-purpose models often struggle with. AlphaGeometry \cite{AlphaGeometry} is another domain-specific model in mathematics that specializes in solving Olympiad-level geometry problems by combining supervised learning and reinforcement learning (RL). In the medical domain, MedPaLM-2 \cite{MedPaLM-2} has been developed to process and interpret medical texts, clinical notes, and health records with a high level of accuracy and relevance. Similarly, ChatLaw \cite{ChatLaw} is finetuned on legal documents and law cases in order to effectively perform legal reasoning.

LLMs have recently attracted significant interest in the telecom industry for their potential to address complex domain-specific tasks. They have shown impressive capabilities in the fault chain tracing for identifying faulty components in a network \cite{00070HCGYBZYSWY23}, and in the classification of technical documents \cite{BariahZZMBD23}. Moreover, LLMs have also been explored in the context of 6G networks, where they support the dynamic selection, deployment, adaptation, and creation of network architectures \cite{abs-2311-05842}. This aligns closely with the emerging concept of AI interconnect \cite{abs-2311-05842}, which focuses on optimizing AI-driven operations across network components. In addition, LLMs have been proposed for automating network configuration generation \cite{abs-2309-06342,MondalTBMV23}, and facilitating FPGA-based wireless hardware development \cite{du2024powerlargelanguagemodels}. Finally, telecom-related question answering (QA) and code generation have been addressed through telecom-related LLMs \cite{TelecomGPT,TelecomRAG,Telco-RAG}. While these tasks are valuable, they are often general in nature and can be handled reasonably well by general-purpose LLMs which already have a broad understanding of telecom-related concepts. However, telecom-related ticket troubleshooting requires deep domain-specific knowledge that is not available in public datasets.

\section{Problem Statement}
 The full ticket troubleshooting process is composed of routing tickets to the appropriate expert team responsible for resolving the ticket (classification task), retrieving contextually and semantically similar historical tickets (retrieval task), and generating a detailed fault analysis report outlining the issue, root cause, and potential solutions (generation task). When training the ticket troubleshooting models of TeleDoCTR, two datasets are given:
 \begin{itemize}
    \item the first dataset $D_t = \{(T_1,t_1),(T_2,t_2),\ldots,(T_{|D_t|},t_{|D_t|})\}$ contains $|D_t|$ ticket-team pairs, where $T_i$ is a ticket and $t_i \in \mathcal{C}$ is the team label from the labels set $\mathcal{C}$.
    \item the second dataset $D_f = \{(T_1,f_1),(T_2,f_2),\ldots,(T_{|D_f|},f_{|D_f|})\}$ contains $|D_f|$ ticket-fault analysis pairs, where $T_i$ is a ticket and $f_i$ is a fault analysis.
\end{itemize}

Each ticket $T_i$ is composed of (1) several attributes $A_i$ such as product name, software and hardware build, title, and problem description; and (2) log snippets $L_i$ that are selected from attached log files to the ticket. Software logs are used in many diagnosis and automation tasks, because they are often the main source of information that is available as a record of information from software runtime. The collected logs are used in multiple log mining tasks such as anomaly detection \cite{loggpt}, failure prediction \cite{ChenYLZGXDZDXLK19}, and failure diagnosis \cite{JiaCYLMX17}. However, a large amount of irrelevant information is typically present in each log file, and this hides the relevant signal that can be used in log mining downstream tasks. Human-assisted log lines selection tools are used to process the collected log files and output a reduced set of relevant log snippets $L_i$. Formally, each ticket $T_i$ is defined as follows:
\begin{equation}
%\small
T_i = A_i \oplus L_i
\label{ticket_info}
\end{equation}
where $\oplus$ denotes the string concatenation operation.

Each fault analysis $f_i$ is divided into three parts: identification $I_i$, which includes the debugging process of the trouble; root cause $RC_i$, a classification from a given root cause set; and resolution $R_i$, which consists of possible solutions for the trouble. Formally, each fault analysis $f_i$ is defined as follows:
\begin{equation}
%\small
f_i = I_i \oplus RC_i \oplus R_i
\end{equation}
After training, TeleDoCTR models are used to solve a given unseen ticket $T_u$ by predicting the team that should resolve the ticket, retrieving similar historical tickets, and generating fault analysis reports detailing how the problem should be resolved.

\begin{figure*}[t!]
\centering
\includegraphics[scale=0.57]{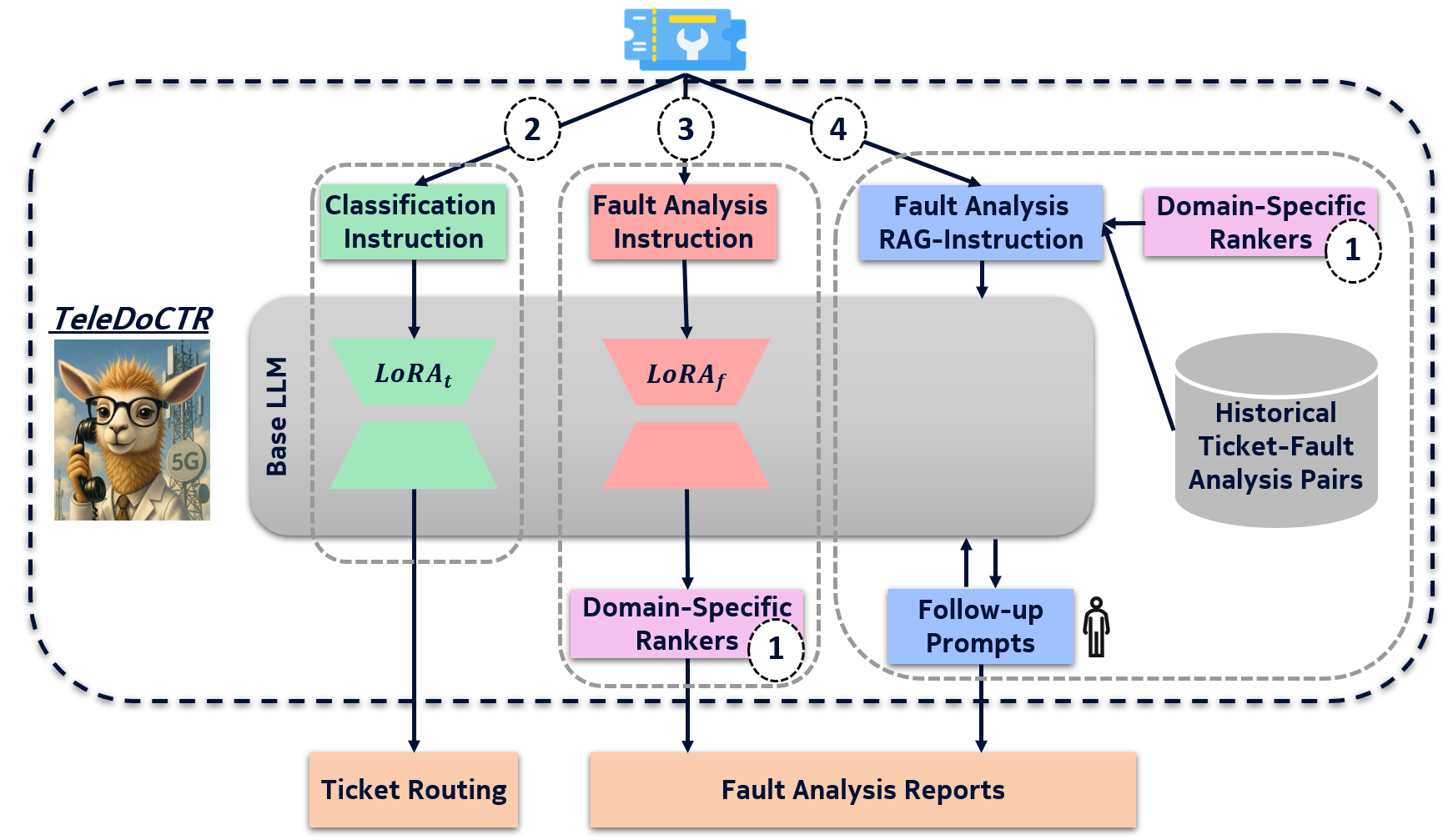}
%\small
\vspace*{-2mm}
\caption{The overview of TeleDoCTR, which has four key components: (1) domain-specific rankers (ranking); (2) a finetuned generative model for ticket routing (classification); (3) a finetuned generative model with multiple fault analysis generation and ranking (generation); and (4) an enhanced RAG-based fault analysis generation with demonstrations selection (generation).}
\label{trs_architecture}
\end{figure*}

\section{TeleDoCTR: Ticket Troubleshooting System}

In this section, we introduce our ticket troubleshooting system TeleDoCTR that takes as input a ticket, and automate key steps of the troubleshooting workflow consisting of: routing tickets to the appropriate expert team (classification task), retrieving contextually and semantically similar historical tickets (retrieval task), and generating a detailed fault analysis report outlining the issue, root cause, and potential solutions (generation task). As shown in Figure \ref{trs_architecture}, TeleDoCTR is composed of four key components: (1) domain-specific rankers; (2) a finetuned generative model for ticket routing; (3) a finetuned generative model with multiple fault analysis generation and ranking; and (4) an enhanced RAG-based fault analysis generation with demonstrations selection. Our domain-specific generative models for troubleshooting are based on a multi-LoRA approach, with a dedicated adapter for each task, while utilizing the same base LLM, denoted as $M$. The use of a shared model with multiple task-specific adapters enables a single deployment, where the appropriate adapter can be dynamically activated based on the selected task. Therefore, TeleDoCTR allows efficient training and inference of the full troubleshooting process. In the next sections, we explain how each of these four components are trained within the context of telecom-related troubleshooting.

\subsection{Domain-Specific Rankers} \label{training_rankers}

Evaluating the similarity between tickets is an important part of ticket troubleshooting to aid in problem resolution. Historical tickets contain valuable information regarding how previous tickets were resolved, which can significantly streamline the full troubleshooting process for new tickets. Similarity between tickets can be defined in various ways such as presence of some common keywords, similarity in some of the ticket attributes, etc. In our model, we capture a more relevant similarity signal to troubleshooting which we define as the fault analysis similarity. In other words, in our similarity definition, two tickets are denoted as similar if they have a similar fault analysis. To this end, we compose two types of pairs from the dataset $D_f$ for training the domain-specific ranker, denoted by $R$. A fault analysis $f_i$ can be associated with multiple tickets $\mathcal{T}_i = \{T_i^j, j=1,2,\ldots,n_i\}$, as these tickets share similar symptoms therefore they are attached to the same fault analysis by domain experts. The first form of pairs is related to implicit ticket similarity captured by ticket-fault analysis pairs composed from $\mathcal{T}_i$ and $f_i$, as $IS_i = \{(T_i^j,f_i), j=1,2,\ldots,n_i\}$. The full ticket-fault analysis data that is composed from all unique fault analyses is equivalent to $D_f$ ($\bigcup_{i=1}^{n} IS_i = D_f$, where $n$ is the total number of unique fault analyses). The second form of pairs is related to explicit ticket similarity captured by ticket-ticket pairs composed from $\mathcal{T}_i$, as $ES_i = \{(T_i^j,T_i^k); j<k; j,k=1,2,\ldots,n_i\}$. The number of pairs in $ES_i$ is $|ES_i| = \dfrac{n_i(n_i-1)}{2}$. The full ticket-ticket data that is composed from all unique fault analyses is defined as $\mathcal{ES} = \bigcup_{i=1}^{n} ES_i$. The  final domain-specific data that is used to train the ranking model $R$ is defined as follows:
\begin{equation}
%\small
\mathcal{S} = D_f \cup \mathcal{ES}
\end{equation}
Tickets and fault analyses can be considered as different text-based modalities. In general, the alignment of modalities is well studied in the literature and benefits from the transformer-based architectures such as in CLIP \cite{clip}, Florence \cite{florence}, ALIGN \cite{align}, MPNet \cite{mpnet}, TSLM \cite{tslm}, etc. In our domain-specific ranker, the objective of the training is to project the embeddings of each modality (ticket modality denoted by $\mathcal{T}_m$ and fault analysis modality denoted by $\mathcal{FA}_m$) into a joint embedding space to enhance the understanding of the relationship between modalities and ensure that the representations are coherent and consistent. This enables us to assess the similarity between modalities that are projected into the shared embedding space (ticket-ticket and ticket-fault analysis similarities). 

The embedding of the ticket $T$ is given by $\boldsymbol{T}=R(T) \in \mathbb{R}^{d}$ and the embedding of the fault analysis $f$ is given by $\boldsymbol{f}=R(f) \in \mathbb{R}^{d}$. The goal is to create an embedding space such that relevant ticket-ticket and ticket-fault analysis pairs will have higher similarity than the irrelevant ones. To train the text-based cross-modal dense retrieval model $R$ using $\mathcal{S}$, we need to create positive and negative pairs. For a given pair $(T,O)$, with $O \in \{\mathcal{T}_m,\mathcal{FA}_m\}$, we denote the negative instances by $O_1^-,O_2^-,\ldots,O_{ng}^-$, where $O_j^- \in \{\mathcal{T}_m,\mathcal{FA}_m\}$ and $ng$ is the number of negatives. We update the parameters of the dense retrieval model $R$ by minimizing the log likelihood loss of the relevant (positive) instance $O$:
\begin{equation}
%\small
\begin{aligned}
& L\left(T, O, O_1^-, \cdots, O_{ng}^-\right)
= -\log \frac{e^{\frac{\operatorname{sim}\left(T, O\right)}{Temp}}}{e^{\frac{\operatorname{sim}\left(T, O\right)}{Temp}}+\sum_{j=1}^{ng} e^{\frac{\operatorname{sim}\left(T, O_j^-\right)}{Temp}}}
\end{aligned}
\label{ranker_loss}
\end{equation}
where $Temp$ is the temperature value, and $\operatorname{sim}\left(T, O\right)$ denotes the similarity between the ticket $T$ and the instance $O \in \{\mathcal{T}_m,\mathcal{FA}_m\}$ using the cosine similarity of their vectors:
\begin{equation}
\operatorname{sim}(T, O)=\frac{\boldsymbol{T} \cdot \boldsymbol{O}}{\left\lVert T \right\rVert \left\lVert O \right\rVert}
\label{similarity}
\end{equation}
Given a batch that is composed of $B$ pairs from $\mathcal{S}$, in-batch negatives \cite{KarpukhinOMLWEC20} are used to train the cross-modal dense retrieval model. Let $\boldsymbol{T}_B$ and $\boldsymbol{O}_B$ be the embeddings of tickets and instances with dimension $(B\times d)$, respectively, in the batch of size $B$. We compute the similarity scores matrix $\boldsymbol{SIM} = \boldsymbol{T}_B \boldsymbol{{O}_B}^T \in \mathbb{R}^{B \times B}$, where each row corresponds to a ticket with $B$ candidate instances from $\{\mathcal{T}_m,\mathcal{FA}_m\}$ . The instance in the diagonal position represents the groundtruth instance (ticket or fault analysis) and the remaining $B-1$ instances are negatives.

Given the randomness of negative instances in a batch when training $R$, it is possible to train multiple rankers $R_1, R_2, \ldots, R_{rn}$, where $rn$ is the total number of rankers. First, these rankers can capture different ranking signals from training the model with different negative instances, so the aggregated relevance score computed from multiple rankers covers multiple similarity aspects. Second, suppose that the recall@$k$ of a given order $K$ for a retrieved set $S_i$ is equal to 1 for all rankers $R_i$. This means that the groundtruth similar ticket is in $S_i$ for all rankers $R_i$. Then, the groundtruth ticket is in the intersection set $\bigcap_{i=1}^{rn} S_i$. So, the recall@$k$ for the intersection set $\bigcap_{i=1}^{rn} S_i$ is also equal to 1, but with an order $K'$ that is smaller or equal to $K$ ($K' = |\bigcap_{i=1}^{rn} S_i| \leq |S_i|=K; i=1,2, \ldots rn$). Therefore, the intersection of multiple rankers can eliminate items that are not retrieved by all rankers while preserving a recall@k of 1.

After training the text-based cross-modal dense retrieval models using the domain-specific data $\mathcal{S}$, $R_1, R_2, \ldots, R_{rn}$ are incorporated into improving the fault analysis generation of both the finetuned- and RAG-based models with accurate response and ticket ranking.

%\subsection{Domain-Specific Generative Models}

\subsection{Ticket Routing}

Traditionally, ticket routing is a classification task that consists of predicting the team of each ticket after training a machine learning model using multi-label cross-entropy loss function. To enable the efficient use of a shared LLM with multiple task-specific adapters, we cast the classification task into generation by composing an instruction-tuning dataset to finetune an LLM for ticket routing or team label generation. For each $I_j = (T_j,t_j) \in D_t$, we apply the instruction-tuning template that is shown in Figure \ref{du_finetuning} to obtain the instruction-tuning data for team label generation denoted as $INST_{D_t}$. The pretrained base LLM, denoted as $M$, is finetuned with the instruction-tuning team label supervised data, using the next token prediction task with the teacher forcing on the groundtruth team label $t_j$ for each instance in $INST_{D_t}$. For training efficiency in terms of time and memory, we add ticket routing-based LoRA layers, denoted as $LoRA_t$ into $M$. During the training, only $LoRA_t$ parameters are updated based on the cross-entropy loss computed only on the team label part, while the parameters of the base model $M$ are kept frozen. During the inference phase of ticket routing, the $LoRA_t$ layers are activated and operate in conjunction with the frozen base model $M$, enabling the generation of team labels without the overhead of full-model finetuning.

\begin{figure}[t!]
\centering
\begin{dufinetuningbox}
System prompt: You are an expert Ticket Resolution and Troubleshooting Assistant.\\

User prompt: Predict the team that is responsible for solving the given ticket.\\

\hspace{3cm}Ticket = $T_i$\\

Assistant response: $t_i$

\end{dufinetuningbox}
\caption{Instruction-tuning template for finetuning the team label generation model.}
\label{du_finetuning}
\vspace*{-4mm}
\end{figure}

\subsection{Multiple Fault Analysis Generation and Ranking}

We show how to leverage the ranking models $R_1,R_2, \ldots,R_{rn}$ in a response-ranking mechanism where multiple fault analysis reports are generated by the finetuned LLM and ranked using the domain-specific rankers to identify the most relevant reports.

\subsubsection{Finetuning domain-specific LLM for fault analysis generation} Similar to ticket routing, we compose an instruction-tuning dataset to finetune an LLM for fault analysis generation. For each $I_j = (T_j,f_j) \in D_f$, we apply the instruction-tuning template that is shown in Figure \ref{fa_finetuning} to obtain the instruction-tuning data for fault analysis generation denoted as $INST_{D_f}$. The pretrained base LLM $M$ is finetuned with the instruction-tuning fault analysis supervised data, using the next token prediction task with the teacher forcing on the groundtruth fault analysis $f_j$ for each instance in $INST_{D_f}$. Similar to ticket routing, we add fault analysis-based LoRA layers, denoted as $LoRA_f$, into $M$ that are updated during the training, while the parameters of the base model $M$ are kept frozen. During the inference phase of the fault analysis generation, we activate $LoRA_f$ layers which are used in addition to the base model $M$ to generate a detailed and domain-specific fault analysis. This modular design allows for scalable deployment across multiple tasks by simply swapping in the appropriate LoRA adapter.

\begin{figure}[t!]
\centering
\begin{fafinetuningbox}
System prompt: You are an expert Ticket Resolution and Troubleshooting Assistant.\\

User prompt: Analyze the given ticket by identifying and explaining the problems and symptoms, then generate possible root causes and resolutions.\\

\hspace{3cm}Ticket = $T_i$\\

Assistant response: $f_i$

\end{fafinetuningbox}
\caption{Instruction-tuning template for finetuning the fault analysis generation model.}
\label{fa_finetuning}
\vspace*{-4mm}
\end{figure}

\subsubsection{Response-ranking method for fault analysis reports} Finetuning with domain-specific troubleshooting data allows the base LLM $M$ to internalize domain expertise and adapt effectively to the unique characteristics of telecom-related fault analysis generation. Through this process, our finetuned model learns to generate detailed and context-aware fault analysis reports tailored to the intricacies of the telecom domain. However, due to the inherent complexity and ambiguity often present in telecom troubleshooting, the finetuned model may produce multiple plausible fault analysis candidates, each capturing different aspects of the trouble with varying levels of certainty. To address this challenge and ensure the reliability of the generated fault analysis reports, we incorporate our domain-specific rankers $R_1, R_2, \ldots, R_{rn}$ that are described in Section \ref{training_rankers}. These rankers are trained to assess the semantic relevance and contextual alignment of each candidate fault analysis with the given ticket. In general, scoring introduces fewer uncertainties compared to generation. This is because, in the scoring process, the complete fault analysis and the corresponding ticket are both available upfront which allows the model to directly assess their semantic relevance. In contrast, generation is inherently autoregressive which means that the LLM generates the fault analysis token by token, with each step conditioned only on previously generated tokens. This sequential nature introduces uncertainty, as the model does not have access to the full context of the final output during the generation process. Therefore, by computing similarity scores, the rankers enable automatic filtering and prioritize the most relevant and accurate responses. This ranking mechanism reduces noise in the outputs and ensures that users receive only trustworthy fault analysis reports.

Formally, for an unseen ticket $T_u$, we generate multiple fault analysis reports $FR = f_1,f_2,\ldots,f_P$, where $P$ is the total number of reports. For each ranker $R_i$, the ranking score $RS_j^i$ between the ticket $T_u$ and the $j^{th}$ fault analysis report $f_j \in FR$ is given by:
\begin{equation}
%\small
RS_j^i = \operatorname{sim}_i(T_u, f_j)
\end{equation}
where $\operatorname{sim}_i$ is the similarity computed as in Equation (\ref{similarity}) with the embeddings extracted from ranker $R_i$. The ranking score $RS_j^i$ assesses the ticket-fault analysis similarities which is enabled through the shared embedding space of $R_i$ including both the ticket and fault analysis embeddings. The final ranking score $RS_j$ between the unseen ticket $T_u$ and $f_j \in FR$ is given by:
\begin{equation}
%\small
RS_j = \dfrac{\sum_{i=1}^{rn} RS_j^i}{rn}
\label{fa_similarity_eq}
\end{equation}
The final ranking score $RS_j$ is a multi-ranker score that captures multiple similarity signals captured from training multiple rankers. Based on the scores $RS_1, RS_2, \ldots, RS_P$, we return the fault analysis report which has the highest score.

\subsubsection{Model alignment using RL from automated feedback} The previous response-ranking method targets enhancing the fault analysis generation of the finetuned model during the inference phase by leveraging the domain-specific rankers $R_1, R_2, \ldots, R_{rn}$. We also investigate the possibility of aligning the finetuned fault analysis model with the domain-specific rankers preferences during the training phase with an additional RL-based finetuning. In this regard, we propose a RLRF (Reinforcement Learning from Rankers Feedback) for further training the finetuned fault analysis generation model with a preference dataset that is composed based on the rankers scores. We are given an additional set of ticket-fault analysis pairs, denoted as $D_f^{RL}$, similar to $D_f$ but smaller in size. For each ticket $T_i \in D_f^{RL}$, we generate multiple fault analysis reports $FR_i = f_1^i,f_2^i,\ldots,f_Z^i$, where $Z$ is the total number of reports. Then, we compute the relevance score of each fault analysis using Equation (\ref{fa_similarity_eq}), and we obtain the scores $RS_1^i, RS_2^i, \ldots, RS_Z^i$. If two candidate responses exhibit a sufficiently large difference in score by exceeding a threshold $\tau$, we construct a preference pair where the response with the lower score is selected as the \textit{rejected} example. For the \textit{chosen} example, we use either the groundtruth fault analysis $FA_{\text{gt}}$ with probability $p$, or the higher-scoring generated response with probability $1-p$. In this way, we construct the rankers-based preference dataset $P_{RL}$ where each instance is a triple $(T_i,chosen_i,rejected_i)$. 

The finetuned fault analysis generation model must not only learn the complex task of troubleshooting, but also adapt to its irregular, semi-structured output format. This dual challenge leads to some model instability, often resulting in pathological outputs, such as degeneration cases \cite{LiLF0LCWS23}. To address this, we introduce a histogram-based method for detecting degenerate model outputs. We observe that pathological responses, such as those with excessive repetition of characters, tokens, or numbers, exhibit unusually skewed frequency distributions. Our method compares the normalized character histograms of a generated fault analysis with its groundtruth counterpart, and flags a response as pathological if the character-level frequency difference exceeds a threshold $Th$. We use the detected pathological examples as negative samples to augment our automatically generated preference dataset $P_{RL}$ for RLRF finetuning. These examples serve as counterexamples to guide the model towards producing well-structured and high-quality outputs. We further finetune $LoRA_f$ while keeping the base model $M$ frozen with Direct Preference Optimization (DPO) \cite{rafailov2023direct} to obtain $LoRA_{f+RLRF}$.

\subsection{Enhanced RAG-based Fault Analysis Generation with Demonstrations Selection}

We show how to leverage the ranking models $R_1,R_2, \ldots,R_{rn}$ to select demonstrations (ticket-fault analysis pairs), which are incorporated into the RAG conversational module of $M$ for improved multi-turn fault analysis generation.

\subsubsection{Demonstrations Selection} The retrieval model is the most important component of a RAG-based architecture. A retrieval model takes as input a query and indexed documents, and it returns a set of documents that are relevant to the query. Two categories of queries are possible: keyword- and content-based queries. Different from traditional RAG, which extracts relevant passages from a large corpus of documents using a general-purpose keyword-based search, our task is very domain-specific and requires a specialized query, corpus, and retrieval model. We incorporate a domain-specific content-based search in order to retrieve accurate context for the RAG-based model. First, the query is a new unseen ticket $T_u$ that is composed of several attributes $A_u$ and log snippets $L_u$ as in Equation (\ref{ticket_info}). Second, the indexed documents are the domain-specific ticket-fault analysis pairs from $D_f$. Third, the retrieval models are the domain-specific rankers $R_1, R_2, \ldots, R_{rn}$. 

For each ranker $R_i$, the ranking score $RS_j^i$ between the unseen ticket $T_u$ and the $j^{th}$ historical ticket-fault analysis pair $I_j = (T_j,f_j) \in D_f$ is given by:
\begin{equation}
%\small
RS_j^i = \operatorname{sim}_i(T_u, T_j) + \operatorname{sim}_i(T_u, f_j)
\end{equation}
where $\operatorname{sim}_i$ is the similarity computed as in Equation (\ref{similarity}) with the embeddings extracted from ranker $R_i$. The ranking score $RS_j^i$ assesses the ticket-ticket and ticket-fault analysis similarities to obtain more accurate retrieval scores. This dual-level score computation is enabled through the shared embedding space of $R_i$ that includes both the ticket and fault analysis modalities. The final ranking score $RS_j$ between $T_u$ and the $j^{th}$ instance $I_j = (T_j,f_j) \in D_f$ is given by:
\begin{equation}
%\small
RS_j = \dfrac{\sum_{i=1}^{rn} RS_j^i}{2\times rn}
\label{rag_similarity_eq}
\end{equation}
The final ranking score $RS_j$ is a multi-ranker and multi-modality score that includes multiple similarity signals captured from training multiple rankers with multiple modalities. Based on the scores $RS_1, RS_2, \ldots, RS_{|D_f|}$, we return the set of top-$K$ similar instances $SD = \{(T_r,f_r)\}_{r=1}^{K}$, where $K$ is the number of retrieved instances. $SD$ is the set of demonstrations (ticket-fault analysis pairs) that is incorporated into the RAG-based conversational module for improved multi-turn fault analysis generation.

\begin{figure}[t!]
\centering
\begin{generationbox}
System prompt: You are an expert Ticket Resolution and Troubleshooting Assistant.\\

User prompt: You are given these demonstrations, where each demonstration is composed of a trouble ticket and its accurate fault analysis of the ticket:\\

\hspace{1cm} Demonstration 1:

\hspace{2cm} Ticket = $T_1$

\hspace{2cm} Fault Analysis = $f_1$

\hspace{1cm} Demonstration 2:

\hspace{2cm} Ticket = $T_2$

\hspace{2cm} Fault Analysis = $f_2$

\hspace{1cm} $\vdots$\\

\hspace{1cm} Demonstration $K$:

\hspace{2cm} Ticket = $T_K$

\hspace{2cm} Fault Analysis = $f_K$\\

Please generate a very detailed fault analysis of this new given trouble ticket by relating it to the provided demonstrations:\\

Ticket = $T_u$\\

Please think step by step about the fault analysis of the new given trouble ticket!\\

Assistant response: As an expert ticket resolution assistant, here is my step-by-step detailed fault analysis of the new given ticket by relating it to the provided demonstrations:\\

[FAULT ANALYSIS] \\

My detailed chain-of-thoughts starts with examining the similarities in terms of symptoms and resolutions with the provided demonstrations, then I should leverage the fault analysis part of the similar demonstrations to generate the most reasonable fault analysis of the new ticket.

**Similarities**: \textbf{Actual Response}

\end{generationbox}
\caption{Fault analysis generation prompt for the first round of the RAG-based module.}
\label{rag_first_round}
\vspace*{-4mm}
\end{figure}

\subsubsection{RAG-based Fault Analysis Generation} We deactivate both LoRA layers $LoRA_t$ and $LoRA_f$, and use the base LLM $M$ for the RAG-based fault analysis generation to avoid the catastrophic forgetting \cite{abs-2308-08747} of some of the critical general skills (summarization, reasoning, etc.) that result from finetuning on specific tasks. We leave investigating techniques to reduce the catastrophic forgetting and enable the use of finetuned models for RAG-based generation as a future work. We guide the RAG-based conversational module towards generating the fault analysis of the new unseen ticket by manually composing the first round of chat to incorporate the selected demonstrations from the retrieval step.  The system, user, and assistant roles of the first round of chat are defined as shown in Figure \ref{rag_first_round}. We set the system as a ticket resolution and troubleshooting assistant. For an unseen ticket $T_u$, the user prompt first includes the selected demonstrations $S_i$, then instructs the base LLM $M$ to generate a detailed fault analysis for the ticket $T_u$ by relating it to the provided demonstrations. For the assistant response, we hard-code the first tokens of the response to force the model to apply a chain-of-thoughts (CoT) reasoning while examining the similarities in terms of symptoms and resolutions between $T_u$ and the selected demonstrations $SD$. Then, the assistant $M$ generates its actual response as shown in Figure \ref{rag_first_round}. After the first guided round of chat, the user can ask the assistant follow-up questions within the same context that contains the full first round of chat. Some of the follow-up questions include requesting a deeper analysis about the discovered similarities with the demonstrations, more clarifications about the suggested solutions, etc. Troubleshooting telecom-related tickets poses significant challenges due to the complexity of the domain and the need for deep technical expertise. In many cases, multiple potential resolutions may exist, making it difficult to generate a definitive one-shot solution. As a result, relying solely on a single interaction with an LLM often falls short of resolving the issue. To address this, our domain-specific RAG-based module enables iterative interactions, allowing users to refine, clarify, and expand the LLM-generated response based on follow-up queries.  

The finetuned- and RAG-based fault analysis reports offer complementary perspectives by capturing fine-grained and coarse-grained aspects of troubleshooting, respectively. The finetuned model excels at generating precise resolution plans by directly learning the mapping from ticket information to fault analysis reports. However, these fine-grained reports may lack broader contextual understanding, particularly insights drawn from historical similar tickets. This is where the RAG-based module proves valuable as it enriches the analysis by retrieving semantically similar past tickets and presenting this coarse-grained context in a coherent and user-friendly manner through multi-turn conversation between the user and LLM. By combining the specificity of the finetuned fault analysis generation module with the contextual depth of the RAG-based module, TeleDoCTR delivers a more holistic and informed fault analysis that not only tackles the immediate issue but also situates it within the broader historical framework.

\section{Evaluation} \label{eval}

\subsection{Telecom Troubleshooting Dataset}

%To train and test TeleDoCTR, we use a telecom-related troubleshooting dataset. 

\subsubsection{Troubleshooting Tickets}

When an issue occurs in telecom components, a ticket $T_i$ is opened to describe the problem at various levels of details. A ticket $T_i$ is composed of these fields:
\begin{itemize}
    \item \textit{Product Name}: There are several telecom-related products such as \textit{5G Radio}, \textit{WCDMA Base Station}, \textit{LTE Base Station}, etc. This attribute indicates where the issue happened. 
    \item \textit{Hardware Unit}: It is a physical entity within the product identified as the source or affected component in the reported trouble.
    \item \textit{Software Build}: It is a designated version of system software that is deployed on telecom equipment. It contains code and configurations that determine the functional behavior of the device. In ticket troubleshooting, the software build is used to identify potential bugs, compatibility issues, or known faults related to a specific release.
    \item \textit{Title}: It is a short natural language sentence that provides an overview description of the issue.
    \item \textit{Problem Description}: It is a long template-based text that contains several fields such as detailed test steps, the expected results of the test steps, the actual observed results, the tester analysis, the fault description and occurrence rate, the customer impact analysis, the used flags, etc.
    \item \textit{Log Snippets}: Multiple types of logs, such as system, runtime, and postmortem, are attached to the ticket. Human-assisted log lines selection tools are used to process the collected log files and output a set of relevant log snippets $L_i$.
\end{itemize}

\subsubsection{Fault Analysis Reports}

While solving a ticket, a fault analysis report is composed to store various steps of resolving the problem. A fault analysis $f_i$ is composed of these fields:
\begin{itemize}
    \item \textit{Identification}:  It is the first prepared part of the fault analysis report, and it is a long template-based text that contains summary of the problem, technical description of the problem, dependencies on configuration, and faulty components and versions.
    \item \textit{Root Cause}: It is determined after the identification part, and it consists of a categorical coarse-grained class such as \textit{implementation error}, \textit{compiler error}, \textit{configuration error}, etc. There are in total 88 possible root cause labels.
    \item \textit{Resolution}: This is the last part of the fault analysis report which is produced by considering the input ticket, and the determined identification and root cause parts. It is a medium length template-based text that contains possible workarounds for ticket resolution, description of corrections, and testing requirements.
\end{itemize}
To train effective domain-specific models for both ranking and fault analysis generation, it is crucial to ensure that the training data contains sufficiently informative and content-rich fault analysis reports. To achieve that, we apply a data curation step that filters out low-information samples. Specifically, given a collection of fault analysis reports $f_i$, we first compute the inverse document frequency (IDF) for each token across the dataset to capture token rarity and informativeness. Then, for each report $f_i$, we compute the informativeness score $IS_i$ by averaging the top-$k$ highest IDF values from the tokens in the report. The underlying intuition is that reports composed primarily of common or template-based tokens will yield lower $IS_i$ values, whereas reports containing domain-specific or rare terms related to troubleshooting details will yield higher $IS_i$ scores. The next step is to apply a strict threshold $Th_f$ to retain only reports that surpass a minimum informativeness level, prioritizing precision in our curated dataset over recall. In our implementation, top-$k$ is equal to 15, and $Th_f$ is equal to 5, which ensures that only the most informative samples are selected. This filtering process leads to a high-quality dataset $D_f$ which is composed of 130,781 pairs spanning multiple years. So, the size of the fault analysis-based instruction data used to finetune $LoRA_f$ is 130,781. The automated informativeness-based filtering is critical for minimizing noise and maximizing learning efficiency, especially in complex and domain-specific tasks such as telecom-based troubleshooting. From $f_i \in D_f$ that are associated with multiple tickets, we form the ticket-ticket pairs dataset denoted as $\mathcal{ES}$. The size of $\mathcal{ES}$ is 71,570, and therefore the size of the domain-specific data $\mathcal{S} = D_f \cup \mathcal{ES}$, that is used to train the domain-specific rankers $R_1, R_2, \ldots, R_{rn}$, is 202,351.

\begin{figure}[t!]
\centering
\includegraphics[scale=0.55]{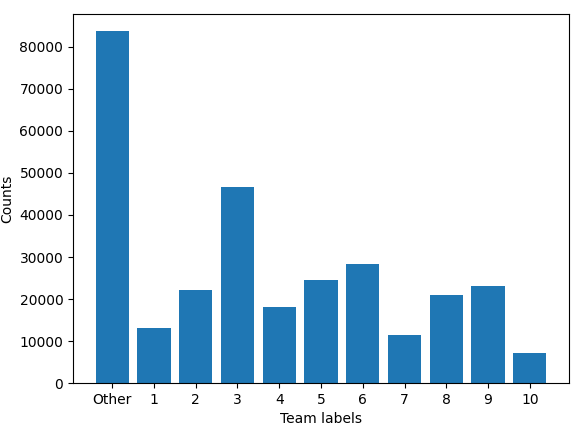}
%\small
\vspace*{-5mm}
\caption{Team labels distribution.}
\label{labels_dist}
\end{figure}

\subsubsection{Team Labels}
Based on the ticket information, there is a team that is assigned to resolve each ticket. The team labels set $\mathcal{C}$ is composed of 11 labels: 10 main teams and an additional label denoted as \textit{Other} that is composed of tickets from the rest of teams. Given the confidentiality of teams, we cannot provide their names. The dataset $D_t$ is composed of 299,495 ticket-team pairs that is used to finetune the ticket routing layers $LoRA_t$. As shown in Figure \ref{labels_dist}, the distribution of tickets is unbalanced.

\subsection{Experimental Setup}

\subsubsection{Domain-specific Rankers}
We set the total number of rankers $rn$ to 6. We initialize each ranker $R_i$ from MPNet\footnote{https://huggingface.co/sentence-transformers/all-mpnet-base-v2} with a dimension $d$ equals to 768. This pretrained model is suitable for our case as it is already trained on a large number of sentence pairs for semantic similarity. In addition, MPNet is a lightweight model that offers efficiency in terms of time and memory for both training and inference phases. A large batch size is key for training an effective ranker that captures multiple aspects of similarities and dissimilarities. Therefore, we set the batch size $B$ to 1024. For the ranking loss computation, we aggregate both the cross entropy of $\boldsymbol{SIM} = \boldsymbol{T}_B \boldsymbol{{O}_B}^T \in \mathbb{R}^{B \times B}$ and $\boldsymbol{SIM}_{sym} = \boldsymbol{O}_B \boldsymbol{{T}_B}^T \in \mathbb{R}^{B \times B}$ to obtain a symmetric loss that captures positive and negative pairs in both directions. Given the large batch size, it is unfeasible to compute the embeddings $\boldsymbol{O}_B$ and $\boldsymbol{{T}_B}$ from a single forward pass. GradCache \cite{GaoZHC21} offers an efficient solution to the memory bottleneck encountered during training with a large batch size. The memory is reduced by decomposing the computation into two stages: an embedding phase and a loss calculation phase, that can be scaled by mini-batches. As a result, memory of constant size (mini batch size = 100 in our case) can now process much larger batches ($B$ = 1024). For the ranker temperature $Temp$, we set it to 0.1 to scale the exponential part in Equation (\ref{ranker_loss}) 10 times. The size of training, validation, and testing sets are 145,693; 16,188; and 40,470 respectively. The ranker is trained with distributed data parallel (DDP) using the accelerator library on 3 $\times$ H100 GPUs. We train the model for 15 epochs and we save the epoch with the lowest loss on the validation set.

\subsubsection{Domain-specific Generative Models}

We set the base pretrained LLM model $M$ as LLaMA-3-8B-Instruct\footnote{https://huggingface.co/meta-llama/Meta-Llama-3-8B-Instruct} for efficient finetuning. We leave finetuning other open-source and lightweight LLMs on the telecom-related troubleshooting data as a future work. We highlight the QLoRA instruction-tuning hyperparameters for both ticket routing and fault analysis generation in Table \ref{finetuning_hyperparams}. Both generative models are finetuned with DDP using the accelerator library on 5 $\times$ H100 GPUs, which means that the actual batch size is 40 after considering the gradient accumulation steps in Table \ref{finetuning_hyperparams}.

For ticket routing, the size of training, validation, and testing sets are 191,669; 47,922; and 59,904 respectively. We train the model for 10 epochs and we save the epoch with the lowest loss on the validation set. For ticket routing inference, we use a greedy search strategy with a beam width of 2 to consistently generate the top-2 tokens. These tokens correspond to the top-2 most probable team labels, as determined by the finetuned model’s probability distribution over the LLaMA vocabulary.

\begin{table}[h]
\centering
\caption{Instruction-tuning hyperparameters for both ticket routing and fault analysis generation.}
\begin{tabular}{ll}
\textbf{Hyperparameters}      & \textbf{Value} \\ \hline
Max Sequence Length           & 10,000          \\
Learning Rate                 & 5e-05          \\
Epochs                        & 10             \\
Batch size per GPU            & 1              \\
Gradient Accumulation Steps   & 8              \\
Learning Rate Scheduler       & Cosine         \\
Optimizer                     & AdamW         \\
Loss Function                 & CrossEntropyLoss\\
Training Precision            & bf16             \\
Warm Up ratio                 & 0.1             \\
Quantization Bit              & 4              \\
LoRA Alpha                    & 16             \\
LoRA Target                   & All            \\
LoRA Rank                     & 8              \\
\end{tabular}
\label{finetuning_hyperparams}
\vspace*{-10mm}
\end{table}

For finetuning the domain-specific LLM on the fault analysis generation, we first leave out 1,000 tickets as a testing set for reporting evaluation metrics and comparing against baselines. The remaining ticket-fault analysis pairs are used for finetuning where the size of training and validation sets are 116,803 and 12,978 respectively. We train the model for 10 epochs and we save the epoch with the lowest loss on the validation set. For the response-ranking method for fault analysis generation, for each ticket $T_i$ in the testing set, we generate 5 reports from each temperature value in this range of values (0.1, 0.3, 0.5, 0.7, 0.9), so the total number of reports $P$ is 25. For each generation, top-p \cite{fan_acl} and top-k \cite{Holtzman2020The} sampling are fixed to 0.95 and 50, respectively. For the fault analysis model's alignment using RL from rankers feedback, the total number of generated reports $Z$ is 6 where we generate 1,2, and 3 reports for temperature values 0.1, 0.5, and 0.9 respectively. In this way, the rankers-based preference dataset $P_{RL}$ is composed efficiently. We set the probability $p$ of the selected chosen fault analysis to 0.5. We generate two versions of datasets for $P_{RL}$. The first one $P_{RL}^{1}$ has 5,399 samples and we do not use the detected degenerate LLM responses as rejected. When composing $P_{RL}^{1}$, we set the rankers score difference threshold $\tau$ to 0.3. The second version $P_{RL}^{2}$ uses the pathological examples as negatives and we collect 24,574 instances for DPO finetuning. When composing $P_{RL}^{2}$, we set the frequency difference threshold $Th$ for pathological samples to 0.07.

For the enhanced RAG-based fault analysis generation, we adapt dynamic top-$K$ similar tickets by fixing the budget of tokens for demonstrations (5,000 tokens in our case), and we keep adding demonstrations until we fill our budget of tokens. We set the temperature of the base LLM $M$ for the RAG-based generation to 0.3.

\subsection{Experimental Results}

\begin{table*}
%\vspace*{-5mm}
\caption{Ticket retrieval evaluation results on the testing set of the troubleshooting data.}
\centering
\resizebox{0.95\textwidth}{!}{%
\begin{tabular}{@{}lcccccc@{}}
\toprule
Model & \textbf{Recall@1} &\textbf{Recall@10}&\textbf{Recall@50}&\textbf{Recall@100}&\textbf{Recall@500}&\textbf{Recall@1000} \\ \midrule
BM25   &0.072   &0.113 &0.234 &0.275   &0.371 &0.412\\ 
Vanilla MPNet  &0.000  &0.000 &0.000  &0.001   &0.004 &0.008\\
Single Domain-specific ranker  & 0.200$\pm$0.012 &0.400$\pm$0.017 &0.571$\pm$0.021 & 0.649$\pm$0.012  &0.821$\pm$0.008&0.880$\pm$0.005\\
Single Domain-specific ranker + BM25  & 0.113$\pm$0.015 &0.296$\pm$0.019 &0.540$\pm$0.018 & 0.627$\pm$0.014  &0.810$\pm$0.010 &0.870$\pm$0.005\\\bottomrule
TeleDoCTR (Multiple Domain-specific rankers)   & \textbf{0.251}  & \textbf{0.461} & \textbf{0.627} & \textbf{0.700}   &\textbf{0.853} &\textbf{0.901}\\ \bottomrule
\end{tabular}\label{retrieval_results}}
\end{table*}

\subsubsection{Domain-specific Rankers}
In this part, we study the performance of the domain-specific rankers $R_1, R_2, \ldots, R_{6}$ in TeleDoCTR. We compare our domain-specific rankers against both the lexical and semantic matching signals captured by BM25 \cite{Robertson96okapiat} and vanilla MPNet \cite{mpnet}, respectively. We evaluate the effectiveness of our domain-specific rankers and baseline methods using Recall@$k$, where $k\in \{1,10,50,100,500,1000\}$. For each test instance, Recall@$k$ measures whether the groundtruth ticket appears among the top-$k$ retrieved candidates. If the groundtruth ticket is retrieved within the top-$k$ candidates, a score of 1 is assigned; otherwise, a score of 0 is given. The final Recall@$k$ is computed by averaging these scores over the entire test set. In our setup, retrieval is performed from an index of 130,781 tickets, meaning that Recall@1 corresponds to correctly retrieving the top-1 result out of 130,781 candidates. The reported results are averaged across the held-out test set comprising 40,470 tickets. Table \ref{retrieval_results} shows the performance of different approaches on the testing set of the troubleshooting data. Our results clearly show that the semantic matching signal of the vanilla MPNet model is not suitable for domain-specific retrieval in telecom-related troubleshooting. A pretrained embedding model lacks the specific vocabulary and terminologies of domain-specific troubleshooting, which makes the retrieval inaccurate. We compute the recall@k scores for all 6 domain-specific rankers, and we report the mean score in addition to the standard deviation. Each of the rankers $R_1, R_2, \ldots, R_{6}$ achieves significantly higher Recall@k scores compared to the semantic-based vanilla MPNet model, and to the lexical-based BM25 model. We combine semantic and lexical matching signals in the variation Single domain-specific ranker + BM25, but this does not lead to improving the retrieval scores compared to just using the single domain-specific ranker. This means that in our domain-specific data, the traditional lexical matching signal does not complement the learned domain-specific semantic matching signal. However, when we compute the intersection of all 6 rankers with an order $K$ that equals to 2,500 (Recall@2,500 = 1 for all rankers $R_1, R_2, \ldots, R_{6}$) and combine the ranking scores of tickets in the intersection set, denoted by $\bigcap_{i=1}^{6} S_i$ ($S_i$ is the returned set from the ranker $R_i$ with a size of 2,500 tickets), from all rankers using Equation (\ref{rag_similarity_eq}), we significantly improve the ranking performance compared to each single domain-specific ranker. Notably, the Recall@$K'$ ($K' = |\bigcap_{i=1}^{6} S_i|$) remains 1, indicating that the intersection set  $\bigcap_{i=1}^{6} S_i$ preserves all relevant tickets while discarding those that are not picked by all rankers $R_1, R_2, \ldots, R_{6}$. This filtering step reduces noise and increases the precision of the candidate set. Then, we further refine the intersection set by combining relevance scores from the different rankers, each of which captures different aspects of semantic similarity. This consensus-driven ranking ensures that the highest-ranked tickets reflect strong agreement across multiple rankers which leads to consistent improvements in Recall@$k$ across all evaluated levels.

\begin{table*}[ht]
\caption{Ticket routing evaluation results on the testing set of the troubleshooting data.}
\centering
\begin{tabular}{||c|c|c|c|c|c||}
\hline
\textbf{Category} & \textbf{Method} & \textbf{Accuracy} & \textbf{Macro-P} & \textbf{Macro-R} & \textbf{Macro-F1}  \\ \hline
\multirow{6}{*}{Retrieval} & Vanilla MPNet  & 67.12 &64.50  &61.67 &62.55  \\
& Single Domain-specific ranker  & 75.57  &73.84 &71.85 &72.51 \\
& Multiple Domain-specific rankers  & 75.57  &73.84 &71.85 &72.51 \\
& MPNet finetuned with team labels (ticket only)   & 74.82  &72.54 &70.23 &70.85 \\
& MPNet finetuned with team labels (ticket and fault analysis)  & 76.33  &74.27 &71.79 &72.59 \\
& Joint finetuning of MPNet (team and fault analysis labels)  & 78.59  &76.44 &74.51 &75.18 \\
\hline
\multirow{3}{*}{Classification} & Finetuned BERT   & 77.49  &75.09 &74.12 &74.29 \\
& Pretrained + Finetuned BERT  & 78.94  & 77.31 &75.27  &75.94 \\
& Custom Tokenizer + Pretrained + Finetuned BERT  &  {\ul 79.60}  & {\ul 77.32} & {\ul 76.32}  & {\ul 76.65} \\
\hline
\multirow{1}{*}{Generation} & TeleDoCTR ($LoRA_t$) & \textbf{80.31}  & \textbf{78.43} & \textbf{77.13} & \textbf{77.47} \\ [1ex]
\hline
\end{tabular}
\vspace*{-2mm}
\label{routing_results}
\end{table*}

\subsubsection{Ticket Routing}

In this part, we study the performance of the ticket routing module in TeleDoCTR. We compare three categories of prediction for team label: (1) team label prediction as a retrieval task by leveraging the MPNet model, (2) team label prediction as a classification task by leveraging the BERT model, and (3) team label prediction as a generative task with our PEFT finetuned $LoRA_t$ model. We evaluate the performance of TeleDoCTR and baselines on the ticket routing task using the accuracy, and macro-averaged precision (P), recall (R) and F1-score of predictions on the testing set composed of 59,904 tickets. Table \ref{routing_results} shows the performance of different approaches on the testing set of the troubleshooting data for ticket routing.

For the retrieval category, we have vanilla MPNet, single domain-specific ranker that is finetuned using in-batch negatives loss, and multiple domain-specific rankers as in ticket retrieval. In addition, we have three additional variations: two variations that are finetuned using the team labeling information with the in-batch all triplet loss \cite{HermansBL17}; and one variation that is jointly finetuned using both in-batch negatives loss and in-batch all triplet loss. The triplet loss aims to ensure that the distance between the anchor and positive is smaller than the distance between the anchor and negative by at least a predefined margin. The in-batch all triplet loss considers all possible combinations of anchor, positive, and negative from the batch, subject to the label constraints. In our case, the label's constraint is the team labeling. The variation, MPNet finetuned with team labels (ticket only), is trained using $D_t$ dataset which is composed of ticket-team pairs. In addition to $D_t$, the variation, MPNet finetuned with team labels (ticket and fault analysis), also considers fault analysis-team pairs that can be deduced from joining $D_f$ and $D_t$. For all retrieval baselines, for a given ticket in the testing set, first we index both the ticket and fault analysis parts of historical tickets, second we retrieve top-10 similar tickets, third we compute the team labels distribution in the returned top-10 tickets, and finally we return the most frequent team label as the predicted label of the given ticket. The variation, MPNet finetuned with the team labels using both ticket and fault analysis information, leads to better evaluation metrics than finetuning MPNet using only ticket information with in-batch all triplet loss and finetuning MPNet using in-batch negatives loss with fault analysis information. The in-batch all triplet loss with team labels captures more team similarity signals compared to the in-batch negatives loss with fault analysis information. This means that there are more coarse-grained similarity signals that can be learned from the team labeling information, than the fine-grained similarity signals that are captured in the fault analysis similarity by the domain-specific rankers of TeleDoCTR. In addition, in the case of team label prediction, combining multiple domain-specific rankers does not help to improve the performance compared to the single domain-specific ranker, as the different fine-grained signals are still lacking the coarse-grained similarity signals that are necessary for ticket routing. Lastly, we jointly finetune MPNet with both (1) team labels (ticket and fault analysis) using the in-batch all triplet loss; and (2) ticket-fault analysis pairs using the in-batch negatives loss. In this variation, we capture both coarse- and fine-grained similarities, and this leads to the best team routing results in the retrieval category. 

For the classification category, we have three variations. The first variation is based on finetuning an encoder-only BERT\footnote{https://huggingface.co/distilbert/distilbert-base-uncased} model on $D_t$. We pool the final hidden state $h_{\theta}$ of the first token [CLS] as the representation of the whole input ticket, where $\theta$ denotes the parameters of BERT. Then, a softmax layer, with parameters $W$, is added on top of BERT to predict the probability of a given team label $l$: $p(l \mid h_{\theta})=\operatorname{softmax}(W h_{\theta})$. To inject the telecom-related knowledge into the encoder-only BERT, the second variation has a pretraining phase before the finetuning phase on the ticket routing task, where we pretrain the BERT model with a large number of unlabeled tickets (489,890 tickets) using the Masked Language Modeling (MLM) task \cite{DevlinCLT19}. To further integrate telecom-specific terminologies into BERT, the third variation uses a domain-specific custom tokenizer constructed from the troubleshooting data. After tokenizing the ticket corpus with the standard BERT tokenizer, we identified 1,932,532 unique tokens. The token frequency follows a Zipfian distribution, where a small subset of tokens occurs frequently, while the majority appears rarely. To balance coverage with efficiency, we retain only the top 30,000 telecom-related troubleshooting tokens that are absent from the original BERT vocabulary which contains 30,522 tokens. This augmentation results in a new tokenizer containing 60,522 tokens in total. Next, we pretrain BERT with the updated tokenizer, enabling the model to learn contextual embeddings for the newly added domain-specific tokens via the MLM objective. This pretraining step ensures that the added telecom-related vocabulary is meaningfully integrated into the representation space rather than treated as out-of-vocabulary tokens. At the end of the pretraining phase, the model achieves a validation perplexity of 1.88, indicating effective adaptation to telecom-related troubleshooting data. Finally, we finetune the pretrained BERT with the custom tokenizer on the downstream task of ticket routing. Table \ref{routing_results} shows that the variation Custom Tokenizer + Pretrained + Finetuned BERT achieves the highest performance from the classification category.

TeleDoCTR ($LoRA_t$) demonstrates a slight but consistent improvement over the strongest retrieval- and classification-based baselines for all evaluation metrics. This gain can be attributed to its ability to handle greater model complexity and process longer ticket inputs (up to 10,000 tokens, as shown in Table \ref{finetuning_hyperparams}). Beyond its performance advantage, TeleDoCTR ($LoRA_t$) also introduces significant advantages for real-world deployment. Its design, which integrates LoRA adapters, enables efficient task switching within a single and unified troubleshooting pipeline, rather than relying on multiple fragmented models. This architecture simplifies system maintenance and ensures smoother scalability as new troubleshooting tasks are added. Together, these features make TeleDoCTR ($LoRA_t$) a more sustainable and deployment-ready solution for end-to-end telecom troubleshooting.

\begin{figure}[t!]
\centering
\includegraphics[scale=0.53]{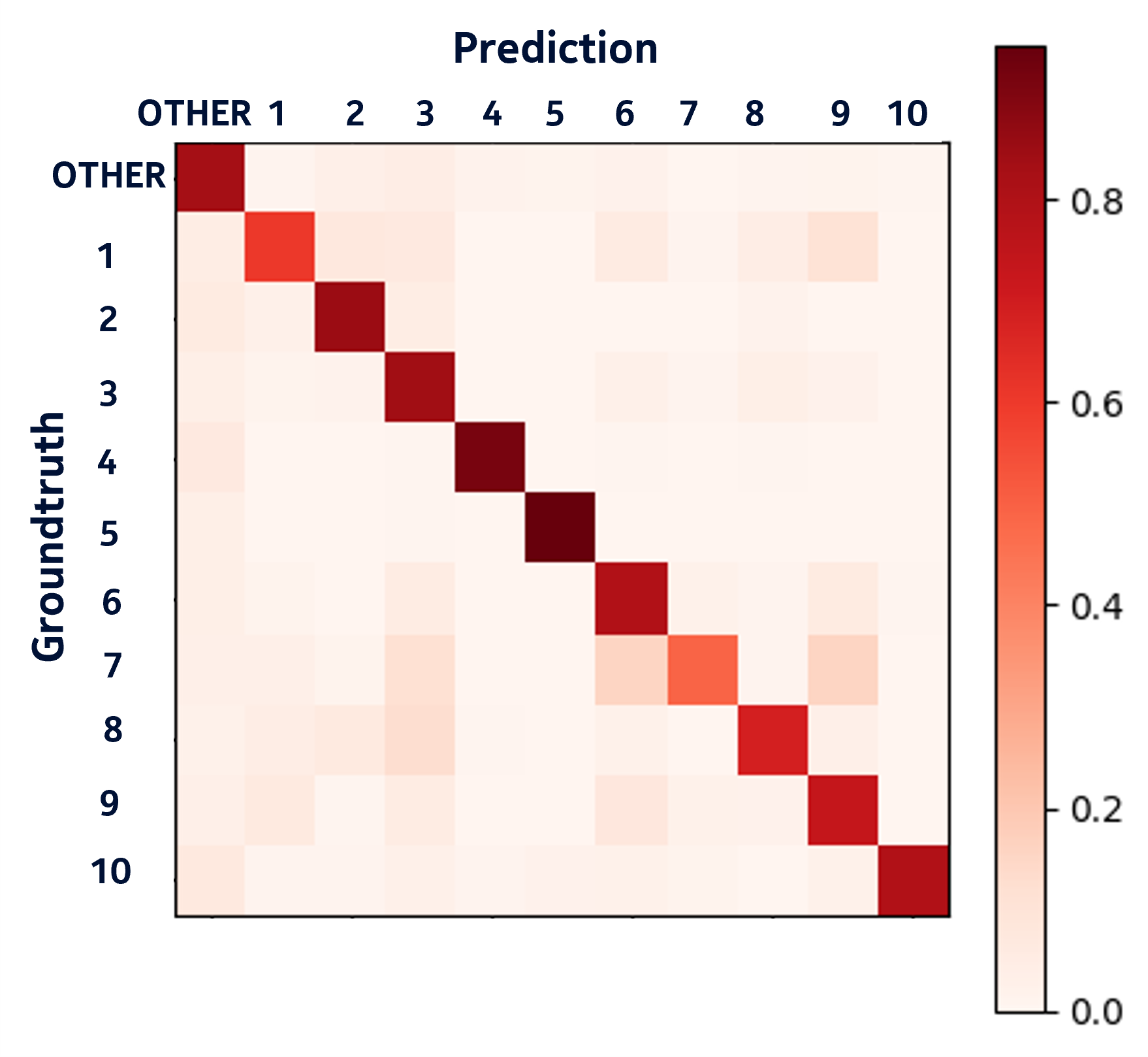}
%\small
\vspace*{-4mm}
\caption{Confusion matrix of TeleDoCTR ($LoRA_t$) for the testing set of ticket routing.}
\label{confusion_matrix}
\end{figure}

In Figure \ref{confusion_matrix}, we show the confusion matrix of TeleDoCTR ($LoRA_t$) for the testing set of ticket routing. In general, the diagonal part of the confusion matrix is clear and reflects the accurate classification results. Based on the diagonal color, the performance ranking of labels from high to low is as follows: 5, 4, OTHER, 10, 3, 2, 6, 8, 9, 1, and 7. Team label 3 is particularly noteworthy, as it represents one of the most challenging cases for accurate prediction. A key objective of our work is therefore to achieve a high F1 score on this label, highlighting the model’s ability to handle difficult routing scenarios. TeleDoCTR ($LoRA_t$) achieves an F1 score of 80.0 for team label 3, representing a substantial improvement over all tested baselines for this specific routing label. This result demonstrates the effectiveness of our approach in addressing challenging ticket routing scenarios and highlights TeleDoCTR as an important milestone toward more accurate and reliable automated ticket routing.

\begin{table*}[ht]
\caption{Fault analysis generation evaluation results on the testing set of the troubleshooting data.}
\centering
\begin{tabular}{||c|c|c|c|c|c||}
\hline
\textbf{Category} & \textbf{Method} & \textbf{ROUGE-1} & \textbf{ROUGE-2} & \textbf{ROUGE-L} & \textbf{BERTScore}  \\ \hline
\multirow{6}{*}{Vanilla} & LLaMA3-8B  &0.216 &0.030  &0.106 &0.800  \\
& LLaMA3-70B  &0.219  &0.030 &0.108 &0.800 \\
& LLaMA3.1-8B  &0.190  &0.021 &0.084 &0.794 \\
& LLaMA3.1-70B  &0.182  &0.020 &0.082 &0.793 \\
& LLaMA3.2-3B   &0.179  &0.017 &0.079 &0.792 \\
& LLaMA3.3-70B  &0.204  &0.026 &0.094 & 0.797\\
\hline
\multirow{3}{*}{RAG} & Vanilla MPNet + General Telecom Documents   &0.218   &0.032 &0.112 &0.806 \\
& TeleDoCTR (Demonstrations Selection with single round)  & 0.266  &0.086 & 0.168 & 0.813\\
& TeleDoCTR (Demonstrations Selection with multiple rounds)  & {\ul 0.371} & {\ul 0.235} &{\ul 0.284}  & {\ul 0.845} \\
\hline
\multirow{3}{*}{Finetuning} & $LoRA_f$ &0.325   &0.191 &0.246 &0.832 \\ 
& $LoRA_f$ + response-ranking with vanilla MPNet  & 0.351  & 0.212& 0.270 & 0.840\\
& TeleDoCTR ($LoRA_f$ + response-ranking with $R_1, R_2, \ldots, R_{6}$)  & \textbf{0.392}   & \textbf{0.252} & \textbf{0.309} & \textbf{0.848}\\[1ex]
\hline
\end{tabular}
\vspace*{-2mm}
\label{fa_generation_results}
\end{table*}

\subsubsection{Fault Analysis Generation}
In this part, we study the performance of the fault analysis generation modules in TeleDoCTR.

\textbf{Quantitative results of fault analysis generation:} We compare three categories of generation for fault analysis: vanilla, RAG, and finetuning. We evaluate the performance of the different methods using three ROUGE \cite{lin2004rouge} scores: ROUGE-1, ROUGE-2, and ROUGE-L; and BERTScore \cite{bertscore}. Table \ref{fa_generation_results} shows the performance of different approaches on the testing set of the troubleshooting data for fault analysis generation. For the vanilla category, we report results using multiple LLaMA models with different scales (ranging from 3B to 70B). All these LLaMA models are the instruction-tuned variants that are optimized for addressing user instructions. For the RAG category, the first baseline (Vanilla MPNet + General Telecom Documents) is a standard RAG approach that uses a vanilla MPNet as a retrieval model and general internal documents related to telecom as the knowledge corpus. For the RAG-based TeleDoCTR module, we evaluate both the single round and multiple rounds of generation for the fault analysis.  The multiple rounds variation simulates the user interactions with 3 follow-up prompts in addition to the main fault analysis generation prompt that is shown in Figure \ref{rag_first_round}. These additional follow-up prompts are: \textit{List the top 3 most similar demonstrations to the new ticket}, \textit{Explain how these identified similar demonstrations were resolved based on their fault analysis reports}, and \textit{Based on all your analysis, provide the final fault analysis report of the new ticket exactly in the following json template of fault analysis by inserting your answers in the indicated sections}. These follow-up prompts further encourage the LLM to double-check the provided demonstrations and explicitly re-rank them to return the top-3 most similar demonstrations and leverage their fault analysis parts for generating a final fault analysis in the required template for the given ticket. For the finetuning category, we compare generating a single fault analysis from the finetuned model ($LoRA_f$) against generating multiple responses and ranking them before returning the most relevant response. For the response-ranking mechanism, we compare ranking responses using vanilla MPNet model ($LoRA_f$ + response-ranking with vanilla MPNet) against ranking using our domain-specific rankers ($LoRA_f$ + response-ranking with $R_1, R_2, \ldots, R_{6}$). Table \ref{fa_generation_results} shows that TeleDoCTR (Demonstrations Selection with single round) significantly outperforms all baselines from the vanilla category in addition to the variant, Vanilla MPNet + General Telecom Documents, from the RAG category. This indicates the importance of incorporating similar historical tickets as a relevant context in RAG using our domain-specific rankers to generate an accurate fault analysis report, as opposed to the standard RAG-based approach of LLMs. Then, the results of the RAG-based TeleDoCTR module are further improved with additional rounds of user interactions for a more explicit refinement of the fault analysis report, as shown in the results of TeleDoCTR (Demonstrations selection with multiple rounds). For the finetuning category, Table \ref{fa_generation_results} shows that if we increase the test-time compute by allowing the finetuned LLM to generate multiple reports for the ranking mechanism (both MPNet and $R_1, R_2, \ldots, R_{6}$ models), we further improve the quantitative metrics, compared to generating a single report after finetuning ($LoRA_f$). We achieve the best quantitative metrics when we rank the responses from the finetuned model using our domain-specific rankers in TeleDoCTR ($LoRA_f$ + response-ranking with $R_1, R_2, \ldots, R_{6}$).

\begin{table*}[ht]
\caption{RLRF evaluation results on the testing set of the troubleshooting data.}
\centering
\resizebox{0.9\textwidth}{!}{%
\begin{tabular}{@{}lcccccc@{}}
\toprule
\multicolumn{1}{c}{\textbf{Dataset}} & \textbf{\begin{tabular}[c]{@{}c@{}}Optimization\\ Steps\end{tabular}} & \textbf{\begin{tabular}[c]{@{}c@{}}Rankers\\ Score$\uparrow$\end{tabular}} & \textbf{\begin{tabular}[c]{@{}c@{}}Pathology\\ Ratio$\downarrow$\end{tabular}} & \textbf{BLUE$\uparrow$}   & \textbf{ROUGE-L$\uparrow$}  & \textbf{METEOR$\uparrow$} \\ \hline
Groundtruth testing set                         & N/A                                                                       & {$\bf{0.60}$}                                        & -                                                                  & -               & -               & -               \\
$D_f$ with ticket-fault analysis pairs                   & $LoRA_f$                                                                       & $0.48$                                                 & {\ul $1.4\%$}                                                      & \textbf{0.29} & \bf{0.31} & \textbf{0.21} \\
Preference Dataset $P_{RL}^{1}$                & $LoRA_{f+RLRF}$                                                                & {\ul $0.53$}                                           & $3.7\%$                                                            & {\ul $0.25$}    & $0.25$          & $0.17$          \\
Preference Dataset $P_{RL}^{2}$                & $LoRA_{f+RLRF}$                                                                & $0.52$                                                 & \textbf{0.7\%}                                                   & \textbf{0.29} & {\ul $0.30$}    & {\ul $0.20$}\\ \bottomrule
\end{tabular}
\label{tab:rlhf-acc}}
\end{table*}

\textbf{RLRF results:} We compare $LoRA_f$ and $LoRA_{f+RLRF}$ using multiple evaluation metrics to understand the contributions of RLRF. For the testing dataset composed of 1,000 tickets, we compute 3 types of metrics. The first type is the text similarity between the generated and groundtruth fault analysis that is captured by BLUE \cite{papineni2002bleu} (it focuses on precision by measuring the overlap of n-grams between generated and reference texts), ROUGE-L \cite{lin2004rouge} (it focuses on recall by measuring how much the reference text is covered by the generated text), and METEOR \cite{banerjee2005meteor} (it combines precision and recall while considering synonyms and stemming). Collectively, these metrics provide a multi-faceted view of lexical and structural similarity which leads to a more fine-grained assessment of the generated fault analysis reports by $LoRA_f$ and $LoRA_{f+RLRF}$. The second type is the domain-specific rankers similarity which is the average over all the testing set of similarities computed using Equation (\ref{fa_similarity_eq}) for each ticket-fault analysis pair in the testing set. This metric measures the alignment between the generated fault analysis and the domain-specific rankers. The third type is the pathology ratio which reflects the pathological responses that result from the degenerate model outputs. Table \ref{tab:rlhf-acc} shows the evaluation metrics of $LoRA_f$ and $LoRA_{f+RLRF}$ on the testing set of the troubleshooting data for fault analysis generation. To estimate the upper-bound performance in terms of domain-specific rankers similarity, we compute the rankers score of the groundtruth ticket-fault analysis pairs of the testing set which is equal to 0.60 as shown in Table \ref{tab:rlhf-acc}. Compared to $LoRA_f$, $LoRA_{f+RLRF}$, that is finetuned using the preference dataset $P_{RL}^{1}$, helps to increase the rankers score and push it closer to the upper bound. However, there are negative effects reflected by the increase in the pathology ratio and decrease in BLUE, ROUGE-L, and METEOR. These negative effects are all reduced with $LoRA_{f+RLRF}$ that is finetuned using the preference dataset $P_{RL}^{2}$ which contains rejected pathological responses. $LoRA_{f+RLRF}$ finetuned with $P_{RL}^{2}$ also maintains a very close rankers score to the model finetuned with $P_{RL}^{1}$. These results show the effectiveness of RLRF in terms of (1) further aligning the finetuned fault analysis generative model with the domain-specific rankers feedback, and (2) reducing the pathological responses that result from degenerate model outputs. Future work includes further improving RLRF to also increase the text-based matching metrics such as BLEU, ROUGE, and METEOR.

\textbf{Qualitative results of fault analysis generation:} We qualitatively evaluate the fault analysis reports generated by the top-2 methods in Table \ref{fa_generation_results} which are: (1) the enhanced RAG module, TeleDoCTR (Demonstrations selection with multiple rounds), and (2) the finetuned module with response-ranking mechanism, TeleDoCTR ($LoRA_f$ + response-ranking with $R_1, R_2, \ldots, R_{6}$). We use LLM as a judge to evaluate four criteria which are: accuracy (To what extent is the predicted fault analysis semantically aligned with the groundtruth fault analysis?), completeness (Does the predicted fault analysis capture all key aspects of the groundtruth fault analysis (main cause, contributing factors, resolution hints)?), relevance (Does the predicted fault analysis stay on the topic of the provided ticket and avoid irrelevant details?), and clarity (Is the reasoning of the predicted fault analysis clear and well-structured?). The LLM takes as inputs the ticket information, the groundtruth fault analysis, and the predicted fault analysis, and predicts a rating score from 0 (poor) to 5 (excellent) for each criterion in addition to a justification for each score. We use LLaMA3.3-70B\footnote{https://huggingface.co/meta-llama/Llama-3.3-70B-Instruct} as the LLM judge and evaluate the predicted fault analyses of the testing data composed of 1,000 tickets. The qualitative evaluation shows that both TeleDoCTR (Demonstrations selection with multiple rounds) and TeleDoCTR ($LoRA_f$ + response-ranking with $R_1, R_2, \ldots, R_{6}$) have mean scores larger than 4 for the relevance and clarity criteria. Specifically, TeleDoCTR (Demonstrations selection with multiple rounds) achieves $4.87\pm0.36$ for relevance and $4.32\pm0.54$ for clarity, while TeleDoCTR ($LoRA_f$ + response-ranking with $R_1, R_2, \ldots, R_{6}$) achieves $4.06\pm0.56$ for relevance and $4.03\pm0.63$ for clarity. However, differences emerge in accuracy and completeness criteria. TeleDoCTR (Demonstrations selection with multiple rounds) achieves $3.33\pm0.76$ for accuracy and $3.32\pm0.84$ for completeness, outperforming TeleDoCTR ($LoRA_f$ + response-ranking with $R_1, R_2, \ldots, R_{6}$) which scores $2.56\pm0.78$ for accuracy and $2.58\pm0.87$ for completeness. This discrepancy is largely attributed to the LLM judge’s strict evaluation of root cause prediction. Since the root cause is represented as a short categorical label, even small deviations from the groundtruth lead to heavy penalties in accuracy and completeness. This categorical root cause contributes minimally to the overall instruction-tuning loss during the finetuning of the fault analysis generation model, as the root cause text is significantly shorter than the identification and resolution fields. Therefore, the finetuned model is more prone to generating mismatched labels despite producing otherwise coherent reports. This is different from TeleDoCTR (Demonstrations selection with multiple rounds) which can pick the exact most relevant root cause from the context built by retrieving the most semantically similar tickets. In contrast, quantitative metrics such as ROUGE scores and BERTScore emphasize the overall semantic similarity of the entire generated fault analysis report with the groundtruth report. Under these metrics, TeleDoCTR ($LoRA_f$ + response-ranking with $R_1, R_2, \ldots, R_{6}$) achieves superior performance as highlighted in Table \ref{fa_generation_results}, which shows its strength in producing globally consistent and well-structured outputs. To conclude, the qualitative evaluation suggests that the root cause prediction should be treated as a separate task, similar to the team label prediction, by introducing new LoRA layers specialized only on the generation of root cause from both the ticket and the identification part of the fault analysis. 

Taken together, these findings suggest that TeleDoCTR ($LoRA_f$ + response-ranking with $R_1, R_2, \ldots, R_{6}$) and TeleDoCTR (Demonstrations selection with multiple rounds) offer complementary strengths. TeleDoCTR ($LoRA_f$ + response-ranking with $R_1, R_2, \ldots, R_{6}$) excels at precisely mapping ticket information into detailed resolution plans, which is reflected in its strong quantitative performance. Yet, these fine-grained reports often lack the broader contextual insights that are necessary to situate an issue within the historical troubleshooting knowledge. TeleDoCTR (Demonstrations selection with multiple rounds) fills this gap by retrieving semantically similar historical tickets and delivering the context through multi-round LLM-user interactions. By combining the specificity of the finetuned fault analysis with the contextual awareness of the RAG-based report, TeleDoCTR provides a holistic troubleshooting framework that not only addresses the immediate trouble but also situates it within the broader context of the telecom system.

\section{Conclusions}
In this paper, we proposed a new telecom-related, domain-specific, and contextual troubleshooting system, denoted by TeleDoCTR, which combines domain-adapted ranking and generative models to automate the telecom-related ticket resolution process. Our system involves a multi-LoRA approach to finetune a domain-specific LLM for both ticket classification and fault analysis generation. Additionally, our system involves finetuning specialized ranking models on the troubleshooting dataset to measure semantic relevance. We introduced a response-ranking mechanism in which multiple fault analysis reports are generated by the finetuned LLM and ranked using the domain-specific rankers to identify the most relevant fault analysis reports. To enhance the RAG-based generation, we leveraged the ranking models to select demonstrations (ticket-fault analysis pairs), which are incorporated into the RAG-based conversational module for improved multi-turn fault analysis generation. To train and evaluate TeleDoCTR, we used large-scale telecom-based troubleshooting data, and demonstrated that our new system outperforms the state-of-the-art baselines in ranking, classification, and generation.

Future work includes (1) integrating TeleDoCTR into an agent-based workflow to automate the continuous collection, curation, and adaptation of troubleshooting data; and (2) enhancing the system’s robustness by both mitigating catastrophic forgetting that results from finetuning and introducing self-reflection capability to enable TeleDoCTR to assess and refine its own outputs.

\subsection*{Acknowledgments}
We thank Gabriel G\'orski, Jakub Kozerski, and Bartlomiej Ruszaj from Nokia Mobile Networks for collecting the telecom-related troubleshooting tickets and logs, and deploying the modules of TeleDoCTR system. We thank Ahmet Akyamac and Jin Cao from Nokia Bell Labs for their valuable feedback. We thank our interns Kalliopi Basioti and Ziyi Liu for their valuable and deep analysis of the troubleshooting data.

\bibliographystyle{ACM-Reference-Format}
\bibliography{main}
\appendix

\end{document}